# A Neural Network Looks at Leonardo's(?) *Salvator Mundi*

Steven J. Frank (steve@art-eye-d.com)
and
Andrea M. Frank (andrea@art-eye-d.com)[1]

*Abstract*: We use convolutional neural networks (CNNs) to analyze authorship questions surrounding the works of Leonardo da Vinci — in particular, *Salvator Mundi*, the world's most expensive painting and among the most controversial. Trained on the works of an artist under study and visually comparable works of other artists, our system can identify likely forgeries and shed light on attribution controversies. Leonardo's few extant paintings test the limits of our system and require corroborative techniques of testing and analysis.

## *Introduction*

The paintings of Leonardo da Vinci (1452-1519) represent a particularly challenging body of work for any attribution effort, human or computational. Exalted as the canonical Renaissance genius and polymath, Leonardo's imagination and drafting skills brought extraordinary success to his many endeavors — from painting, sculpture, and drawing to astronomy, botany, and engineering. His pursuit of perfection ensured the great quality, but also the small quantity, of his finished paintings. Experts have identified fewer than 20 attributable in whole or in large part to him. For the connoisseur or scholar, this narrow body of work severely restricts analysis based on signature stylistic expressions or working methods (Berenson, 1952). For automated analysis using data-hungry convolutional neural networks (CNNs), this paucity of images tests the limits of a "deep learning" methodology.

Our approach to analysis is based on the concept of image entropy, which corresponds roughly to visual diversity. While simple geometric shapes have low image entropy, that of a typical painting is dramatically higher. Our system divides an image into tiled segments and examines the visual entropy of each tile. Only those tiles whose entropies at least match that of the source image are used for training and testing. The benefit of what we call our "Salient Slices" approach (Frank et al., 2020) is two-fold. The tiles — unlike the high-resolution source images they represent — are small enough to be processed by conventional CNNs. Moreover, a single high-resolution image can yield hundreds of usable tiles, making it possible to successfully train a CNN even when the number of source images is limited.

---

[1] Co-Founders, Art Eye-D Associates LLC.

We successfully developed and trained CNN models capable of reliably distinguishing the portraits of Rembrandt Harmenszoon van Rijn (1606-1669) and landscape paintings by Vincent Willem van Gogh (1853-1890) from the work of forgers, students, and close imitators. Leonardo's paintings, however, besides being few in number are of mixed genre and subject to varying degrees of authentication controversy. They are also enormously valuable and often hauntingly beautiful. Grappling with this work revealed capabilities we doubted our system possessed and led us to techniques of data augmentation and handling whose success surprised us.

### *Leonardo's Paintings*

Leonardo's subjects include portraits and a variety of religious subjects. His religious paintings subdivide into several different pictorial genres — intimate representations of Madonna and child, portrait-like representations of John the Baptist (*Saint John the Baptist*, 1513-1516; Louvre) and Christ (*Salvator Mundi*, date and current location unknown), and wider-scale scenes with numerous figures and landscape elements. Just the variety of subject matter posed formidable challenges, because our experience with Rembrandt and van Gogh demonstrated that a model trained in one genre can fail spectacularly in another: our Rembrandt portrait models misclassified his religious scenes and our van Gogh landscape models could not distinguish between a genuine self-portrait and a forgery. To have any chance of success, then, a training set utilizing the few confirmed autograph works of Leonardo would require a comparative set of works diverse not only in artists (to promote generalization beyond the training set) but in genre (to span Leonardo's subject matter) — in other words, a comparative training set far larger than the set of Leonardo paintings. Such deliberate lack of balance risked a bias toward false negatives (Chollet, 2018).

Like Rembrandt, Leonardo ran an extensive studio, employing assistants and teaching students. If anything, the contributions made by these associates to the works of the master is even less well understood than for Rembrandt and potentially more significant in many cases, leading to an entire category of "apocryphal" Leonardo works. Even for works that appear to have a single author, experts routinely question whether that author is Leonardo. Excluding all paintings whose attributions to Leonardo have been credibly questioned would leave fewer than half a dozen images for both training and testing.

Yet a further complication is the current state of some Leonardo works. The most definitive provenance is that of *The Last Supper* (c. 1490s; Convent of Santa Maria delle Grazie, Milan) an enormous mural that began to deteriorate shortly after its completion and which is now far too damaged to serve as a training image. Restoration efforts that have been made over the centuries have sometimes involved significant repainting. The recent and highly publicized controversy surrounding *Salvator Mundi*, the world's most expensive painting, is another case in point. Once presumed to be a later copy of a lost original, the panel was purchased in 2005 and restored by the eminent conservator Dianne Modestini. Although the degree of restoration was considerable, Leonardo's *sfumato* technique is evident throughout the painting.

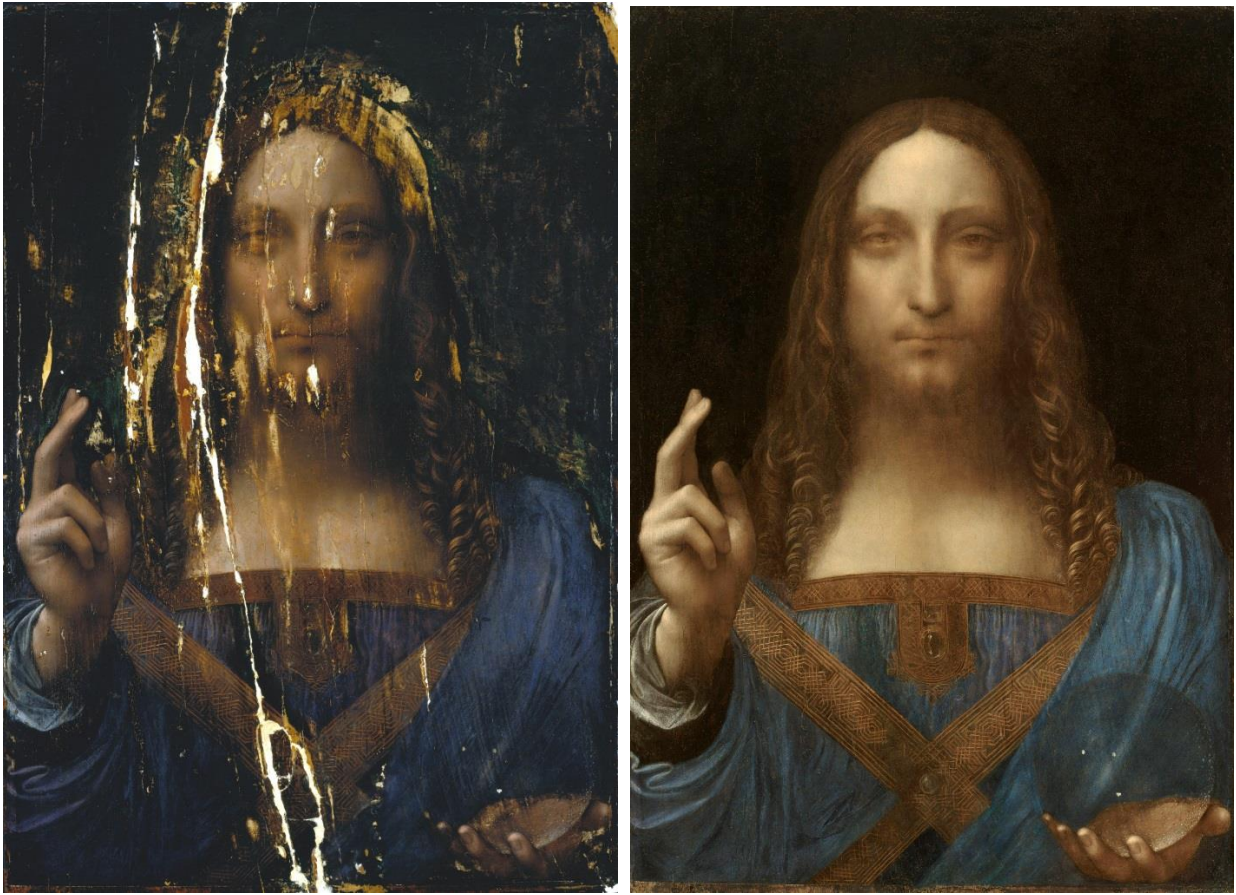

Fig. 1: (a) left, *Salvator Mundi* prior to restoration; (b) right, in 2017 when sold at Christie's, New York, after restoration.

Since then, it has gained some scholarly acceptance as Leonardo's original (Kemp et al., 2020; Syson, 2011) or as partially by Leonardo (Bambach, 2012), while others reject the attribution entirely (Hope, 2020). If we could overcome the considerable technical challenges described above and manage the irreducible authorship uncertainties surrounding Leonardo's work, we might be able to contribute to the discussion as well as explore the effects of restoration on our CNN's performance.

*Methodology*

Our first task would be to assemble all finished paintings at least arguably attributable to Leonardo and assess the strengths of their attributions. Based on this assessment, we would need a strategy for assessing classification accuracy by reserving for testing the smallest possible number of Leonardo works in order to maximize the size of the Leonardo training set. To complete the training set, we would need comparative works by many artists portraying subject matter similar to our Leonardo training images and with varying degrees of pictorial similarity to those images; and somehow, in the end, we would have to wind up with Leonardo and non-Leonardo training tiles roughly equal in number and also sufficiently numerous to support reliable training.

Table 1 below summarizes the works we used, their subject matter, the certainty of attribution, and the use to which we put tiles derived from the image.

| *Title (year)* | *Subject matter* | *Attribution status* | *Use* |
| --- | --- | --- | --- |
| *Mona Lisa* (1503-1506) | Portrait | Substantially unquestioned | Training |
| *The Annunciation* (1472) | Religious scene, multiple figures and landscape | Substantially unquestioned | Training |
| *The Baptism of Christ* (1470-1480) | Portrait (one angel painted by Leonardo) | Generally unquestioned | Training (using isolated Leonardo angel) |
| *Madonna of the Carnation* (1478) | Madonna and child | Substantially unquestioned, possibly with some overpainting | Training |
| *Ginevra de' Benci* (1474-1478) | Portrait | Generally unquestioned | Training |
| *Benois Madonna* (1478) | Madonna and child | Generally unquestioned | Training |
| *Virgin of the Rocks* (Louvre version) (1483-1486) | Religious scene, multiple figures and landscape | Substantially unquestioned | Training |
| *The Lady with an Ermine (Portrait of Cecilia Gallerani)* (1490) | Portrait | Generally unquestioned | Training |
| *The Virgin and Child with Saint Anne* (1503) | Religious scene, multiple figures and landscape | Substantially unquestioned | Training |
| *Saint John the Baptist* (1513-1516) | Religious scene, single figure | Generally unquestioned | Training |
| *Portrait of a Musician* (1490) | Portrait | Generally unquestioned | Training |
| *Virgin of the Rocks* (London version) (1491-1508) | Religious scene, multiple figures and landscape | Generally unquestioned | Training |
| *La Bella Principessa* (1495-1496) | Portrait | Questioned | Test |
| *La Belle Ferronnière* (1490-1497) | Portrait | Generally unquestioned | Test |
| *Madonna Litta* | Madonna and child | Questioned | Comparative |
| *Isleworth Mona Lisa* (1508-1516) | Portrait | Questioned | Comparative |
| *Seated Bacchus* (1510-1515) | Religious (genre) scene, single figure | Workshop of Leonardo | Comparative |

Table 1

We chose *La Belle Ferronnière* as a test image due to its visual similarity to *Lady with an Ermine*, so its absence from the training set would have a smaller impact than sacrificing a more unique image. *La Bella Principessa* may seem an unlikely candidate for a test image: it is a pastel work rather than a painting and its attribution is uncertain. Yet all models that successfully classified both *La Belle Ferronnière* and a large proportion of the non-Leonardo images also invariably classified *La Bella Principessa* as the work of Leonardo with high probability. Interestingly, swapping it for *Portrait of a Musician* in the training set adversely affected model performance. Clearly the strength of a classification does not guarantee that the image will contribute positively to training; the effect of the internal CNN weights on a test image to produce a classification, in other words, is not the same as the influence of the image on the CNN weights during training.

Confining our Leonardo training set to works whose attributions are reasonably secure and hoping somehow to make due with only two test images left us with 12 Leonardo training images — a number that seemed untenably small, particularly compared to the number of comparative (non-Leonardo) training images we ultimately found necessary to produce accurate classifications. Our final comparative training set consisted of 37 images in subject-matter categories corresponding to those listed in Table 1 and in roughly similar proportions. We drew our various training and test sets from a pool of 64 comparative paintings by artists including Leonardo's teacher, Andrea del Verrocchio; his students Giovanni Antonio Boltraffio and Andrea Solario; the Renaissance master Raffaello Sanzio da Urbino (Raphael), who admired and was influenced by Leonardo; unidentified "School of Leonardo" painters; Albrecht Dürer, whose work has been mistaken for Leonardo's; and others including Antonio del Pollaiuolo, Guido Reni, Anna Maria Sirani, Andrea Solari, Georgione, and Giovanni Bellini.

We considered various strategies for boosting the number of Leonardo training tiles and equalizing the number of Leonardo and non-Leonardo tiles. Our first effort, following downsampling of the high-resolution source images to a consistent resolution of 25 pixels/canvas cm, was to isolate heads and faces from the paintings and use an extreme level of tile overlap so that even a single head-size image would yield hundreds of overlapping candidate tiles, which we sifted using our entropy criterion. In particular, our Leonardo tiles (obtained from the heads in our 12-image training set) overlapped by 92% and our non-Leonardo tiles (obtained from the heads of 24 non-Leonardo images) by 88%. Although the difference may seem small, the additional overlap for Leonardo tiles resulted in a twofold increase in their number relative to the non-Leonardo tiles and substantially equalized the populations of Leonardo and non-Leonardo tiles. The price of this data augmentation was significant data redundancy, and the increase in Leonardo tile numbers means that the tiles collectively contained only half the unique information present in the non-Leonardo tiles (which are themselves highly redundant). The effect is exacerbated further by the already small size of a head image, which limits the maximum tile size and, therefore, the amount of visual information that can be analyzed.

Despite our pessimism given these severe data limitations, the models we generated at the maximum usable tile size performed quite well, achieving 94% accuracy. At this preliminary stage, using heads from only two Leonardo test images (and from 13 comparative test images), we considered models producing even a single false negative — i.e., an improperly classified Leonardo — to be failures. But obviously we would need further strategies to validate what could easily represent a misleadingly favorable result; more Leonardo test images, were they available, might reveal those results to be lucky anomalies.

The results did seem to suggest that the sheer quantity of tiles might be more important to classification success than their unique information content. Thus emboldened, we considered using the same approach on the full-size images, which would allow us to test many more candidate tile sizes. Once again we were pessimistic, this time because of the mixed genres. Nonetheless, as we did for Rembrandt and van Gogh, we tested a succession of tile sizes ranging from 100×100 to 650×650 pixels and found peak accuracy for Leonardo to occur at 350×350 pixels — close to the optimal size for Rembrandt. That accuracy was only 82%, unfortunately, but we obtained steady improvement as we increased the size of the comparative training set. Of course that also required a relative increase in the overlap of the Leonardo tiles, and in fact, both Leonardo and non-Leonardo tiles needed more overlap in order to generate sufficient tile populations. We finally achieved equal and sufficient numbers of Leonardo and comparative tiles at overlaps of 94% and 92%, respectively. Because of the two-dimensional geometry involved, the 2% difference in overlap resulted in three times as many tiles per Leonardo image relative to the non-Leonardo images. Using our 12 Leonardo training images and 33 non-Leonardo training images (but substantially similar numbers of Leonardo and non-Leonardo tiles), we obtained an in-sample accuracy of 97% on a test set with 31 non-Leonardo and our two Leonardo test images, with no false negatives.

Now we needed a way to corroborate the results tentatively suggested by a test set severely deficient in Leonardo images. We adopted several expedients. First we shuffled our comparative training and test sets, preparing four new tile sets with randomly selected splits of 32 test images and 32 training images. We trained and tested 350×350 models for each of the new sets. The best-performing models derived from each new set exhibited test accuracies within a relatively narrow band (90-94%) and, as expected, underperformed our curated training set. One of the four sets failed to produce a model free of false-negative classifications, suggesting that successfully classifying our two Leonardo test images while also properly classifying most of the comparative test images (i.e., avoiding false positives) is not trivial.

As an external test, we used our best-performing models from both the curated and random tile sets to classify *Seated Bacchus*, once erroneously attributed to Leonardo, to see whether a painting that had once fooled experts could reveal deficiencies in our (inadequately tested) models. In fact, all successful models — i.e., the ones free of false negatives — strongly classified *Seated Bacchus* as not painted by Leonardo (with the best model derived from the curated set assigning a 100% classification probability). This provides some evidence that our models are not prone to false positives.

We also tested our best-performing models from both the curated and random tile sets on the *Madonna Litta* and the *Isleworth Mona Lisa*, hoping to find consistency among models notwithstanding the different training sets. All successful models classified the *Isleworth Mona Lisa* as not painted by Leonardo.[2] The results were more complex for the *Madonna Litta*. The best models from our curated set and one of the random sets solidly classified this painting as not by Leonardo. The two other successful random sets each yielded two models that, despite identical accuracy scores, classified *Madonna Litta* differently from each other. Nonetheless, in each case, the model that more strongly classified *Seated Bacchus* and the *Isleworth Mona Lisa* as not painted by Leonardo also classified the *Madonna Litta* as not by Leonardo. This behavior — with classification tendencies moving together consistently and progressively — suggests model stability across training sets, which would be expected of any reliable and methodologically sound model.

---

[2] The best image we could obtain for this now-hidden work is of unfortunately poor quality, and image artifacts appear to have distorted our probability maps. While useful to confirm behavior consistency among models, the classifications we obtained for this painting are not otherwise meaningful.

Finally, to further test model stability, we tried altering the architecture of our CNN. In particular, we increased the number of convolutional layers from five to eight, and increased the size of the convolution "kernel" — the CNN's feature extractor — in the early layers. Models based on this eight-layer architecture consistently outperformed their five-layer counterparts, with the best curated-set model achieving 100% classification accuracy and all of the random-set models delivering accuracies of 82% to 97% with no false negatives. Here, the best models derived from the curated set and all random sets very strongly classified *Seated Bacchus*, and solidly classified the *Isleworth Mona Lisa* and the *Madonna Litta*, as not Leonardo works.

### *Results: Who Painted Salvator Mundi?*

This suggestive level of corroboration convinced us that we were ready to analyze *Salvator Mundi*. We used the best five-layer and eight-layer models generated from our curated data set to create the probability maps shown in Figs. 2(a) and 2(b).

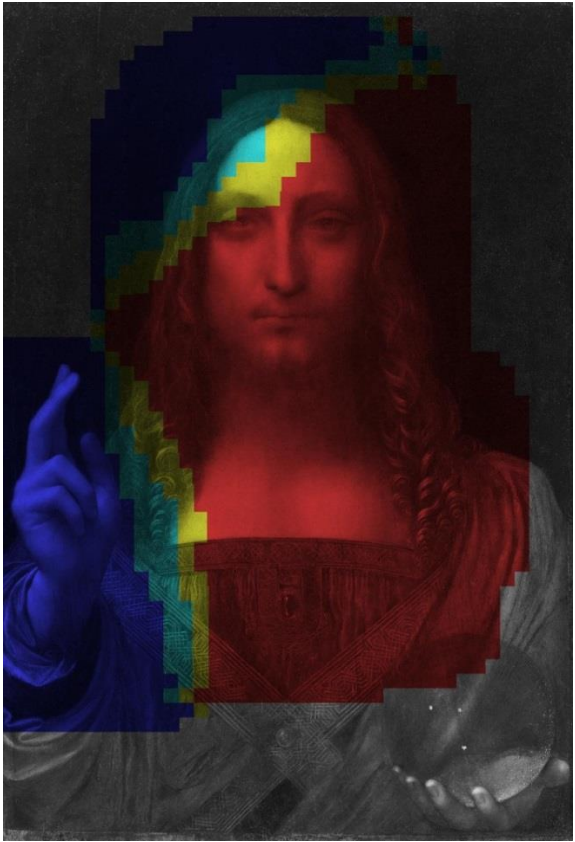
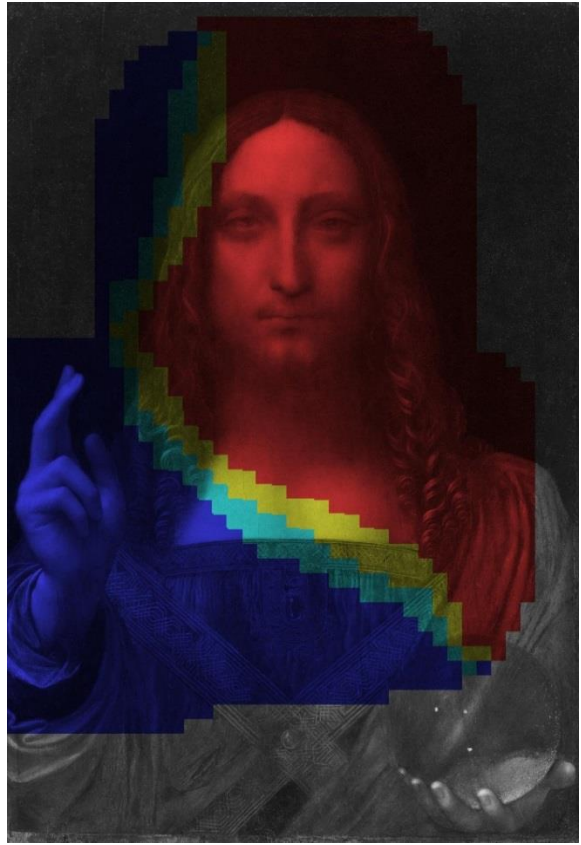
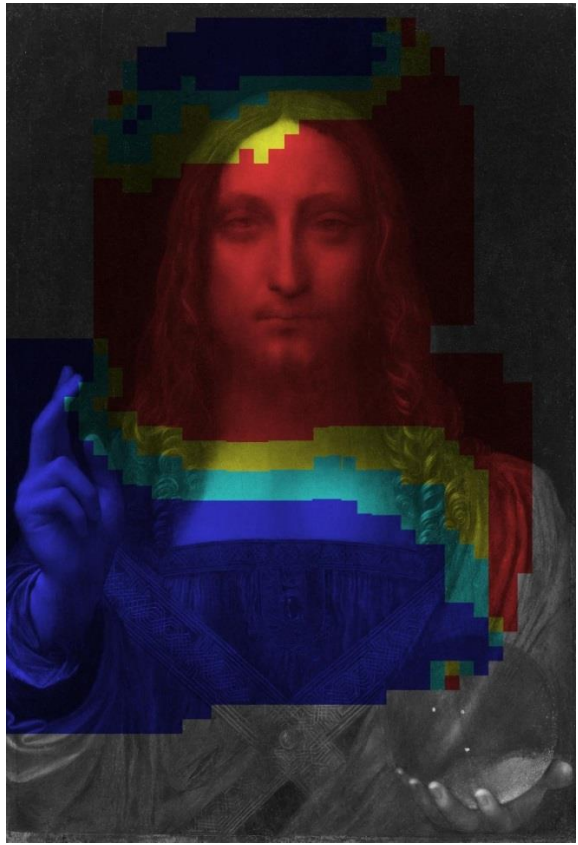

Fig. 2: (a) Top left, probability map for *Salvator Mundi* generated from best eight-layer model trained on the curated dataset. (b) Top right, probability map generated from best five-layer model trained on the curated dataset. The maps color-code the probabilities assigned to the examined regions of an image at a granular level: red corresponds to high-likelihood ($\geq 0.65$) classification as Leonardo, gold to moderate-likelihood ($0.5 \leq p < 0.65$) classification as Leonardo, green to moderate-likelihood ($0.5 > p > 0.35$) classification as not Leonardo, and blue to high-likelihood ($\leq 0.35$) classification as not Leonardo. (c) Bottom, probability map for *Salvator Mundi* generated from best random set

Both maps exhibit largely similar probability distributions, classifying the "blessing" hand and a portion of the background as not painted by Leonardo. One possible explanation for the blue classification of the background and, in the five-layer map, a portion of the chest garment is the degree of damage to (and consequent extensive restoration of) those areas. But in fact, despite considerable damage to the facial region, it is strongly classified as Leonardo in the restored painting. Our maps therefore suggest that the restorer did a magnificent job, and that the most important parts of the painting are indeed Leonardo's work. (The small area of light blue along the hair and forehead in the left-side probability map is likely spurious spillover from the dark blue classification of the adjacent background; this spillover arises from the way probabilities are combined among the fairly large tiles to produce the final map.) The left-hand map, generated by the more accurate model, confines the lower blue portion to the blessing hand. Artists who employed assistants and taught students (Rembrandt, for example) often directed those who could emulate the master's technique to paint "unimportant" elements such as hands, either for efficiency or as an exercise (van de Wetering, 2017). During restoration, a prominent "pentimento" — a change in composition made by the artist in the finished work — was observed in the thumb of the blessing hand (Christie's, 2017).

Indeed, the blessing hand has been the subject of much scholarly controversy. One expert believes that "much of the original surface [of *Salvator Mundi*] may be by Boltraffio, but with passages done by Leonardo himself, namely Christ's proper right blessing hand, portions of the sleeve, his left hand and the crystal orb he holds" (Bambach, 2012). Another argues the opposite: "The flesh tones of the blessing hand, for example, appear pallid and waxen as in a number of workshop paintings. … It is therefore not surprising that a number of reviewers of the London Leonardo exhibition initially adopted a skeptical stance towards the attribution of the New York Salvator Mundi" (Zollner, 2017). Given all of this, the probability distribution given by our most accurate model does not appear to be an unreasonable one.

The overall probabilities assigned to *Salvator Mundi* by the best eight-layer and five-layer model are, respectively, 0.74 and 0.62. What about models generated using the random datasets? The results for eight-layer models are summarized in Table 2.

| Random Set # | Model # | Accuracy | False Negatives | False Positives | Salvator Mundi |
|---|---|---|---|---|---|
| Set 0 | 33 | 0.93 | 0 | 2 | 0.82 |
| Set 1 | 21 | 0.97 | 0 | 1 | 0.55 |
|  | 22 | 0.91 | 0 | 3 | 0.8 |
|  | 24 | 0.91 | 0 | 3 | 0.81 |
| Set 2 | 32 | 0.82 | 0 | 6 | 0.93 |
| Set 3 | 17 | 1 | 0 | 0 | 0.55 |
|  | 20 | 0.97 | 0 | 1 | 0.64 |

Table 2

Strikingly, as illustrated in Fig. 3, there is an almost linear relationship ($R^2 = 0.81$) between the number of false positives produced by a model and the overall probability score that it assigns to *Salvator Mundi*.

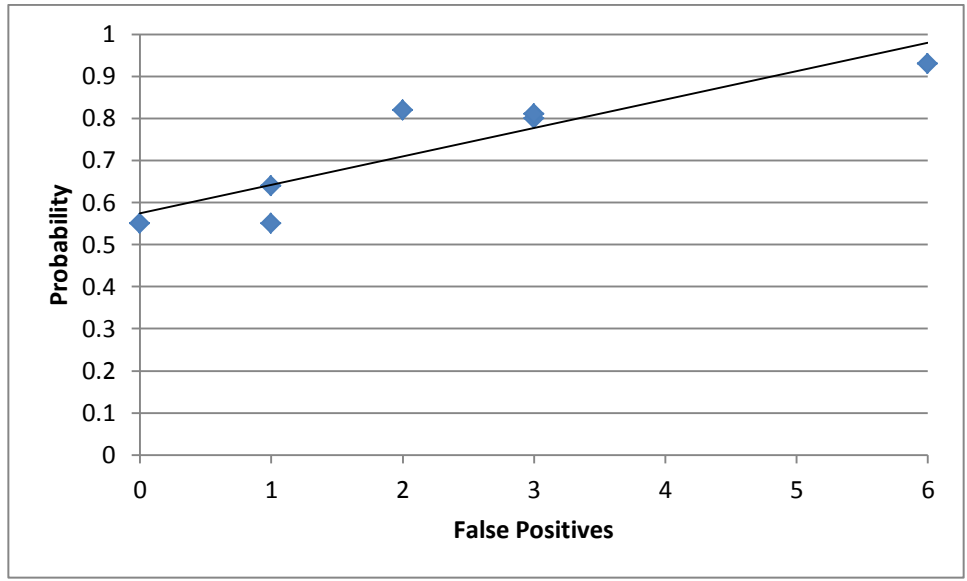

Fig. 3: Overall probability assigned to *Salvator Mundi* by various models as a function of false positives the model produces

In other words, the more lenient a model is in classifying close calls as Leonardos, the more of *Salvator Mundi* the model will classify as painted by Leonardo. The best model from random set #3, which achieved an accuracy of 100%, assigns *Salvator Mundi* an overall probability of 0.55 and produces the probability map shown in Fig. 2(c), which is nearly identical to Fig. 2(b). Because both the training and test sets were generated randomly in this case, we have more confidence in the map of Fig. 2(a), which reflects curatorial efforts to balance the types of work in both training and test sets; a perfect score achieved by a randomly generated set likely has some stochastic (e.g., lucky) origin. But the persistence of the general probability pattern across models generated with different training sets and with different model architectures seems, once again, to offer a measure of cross-validation.

## *Conclusion*

With enough training and test images and curatorial attention to their distribution and character, our Salient Slices technique produces classifications consistent with the current scholarly consensus. Yet even with image bases that appear unmanageably small, high degrees of data augmentation combined with corroborative testing strategies permit meaningful classifications, even at the subimage level. We hope that *Salvator Mundi*, whose present whereabouts are unknown, emerges from hiding and assumes its rightful place in Leonardo's oeuvre.

F. Zollner, Leonardo da Vinci: The Complete Paintings and Drawings. Köln: Taschen, 2017, pp. 440–445.